\documentclass{article}
\usepackage{spconf,amsmath,graphicx,booktabs}

\usepackage{hyperref}
\usepackage{booktabs}
\usepackage{amssymb}
\usepackage{makecell}
\usepackage{xcolor}
\newcommand{\downdata}[1]{\textcolor{blue}{#1}}
\newcommand{\risedata}[1]{\textcolor{red}{#1}}

\title{MultiWay-Adapter: Adapting Multimodal Large Language Models for scalable image-text retrieval}
\name{Zijun Long, George Killick, Richard McCreadie, Gerardo Aragon Camarasa}
\address{The University of Glasgow, Scotland, UK}
\begin{document}
%
\maketitle
\begin{abstract}
As Multimodal Large Language Models (MLLMs) grow in size, adapting them to specialized tasks becomes increasingly challenging due to high computational and memory demands. Indeed, traditional fine-tuning methods are costly, due to the need for extensive, task-specific training. While efficient adaptation methods exist that aim to reduce these costs, in practice they suffer from shallow inter-modal alignment, which severely hurts model effectiveness. To tackle these computational challenges and improve inter-modal alignment, we introduce the MultiWay-Adapter (MWA), a novel framework featuring an `Alignment Enhancer'. This enhancer deepens inter-modal alignment, enabling high transferability with minimal tuning effort. Our experiments show that unlike prior efficient tuning approaches, MWA maintains model effectiveness, while reducing training time by up-to 57\%. MWA is also lightweight, increasing model size by only 2-3\% (in terms of parameters) for state-of-the-art foundation models like BEiT-3 Large. These results demonstrate that MWA provides an efficient and effective adaptation method for MLLMs, significantly broadening their applicability.

\end{abstract}

\begin{figure*}
    \centering
    \includegraphics[width=14cm]{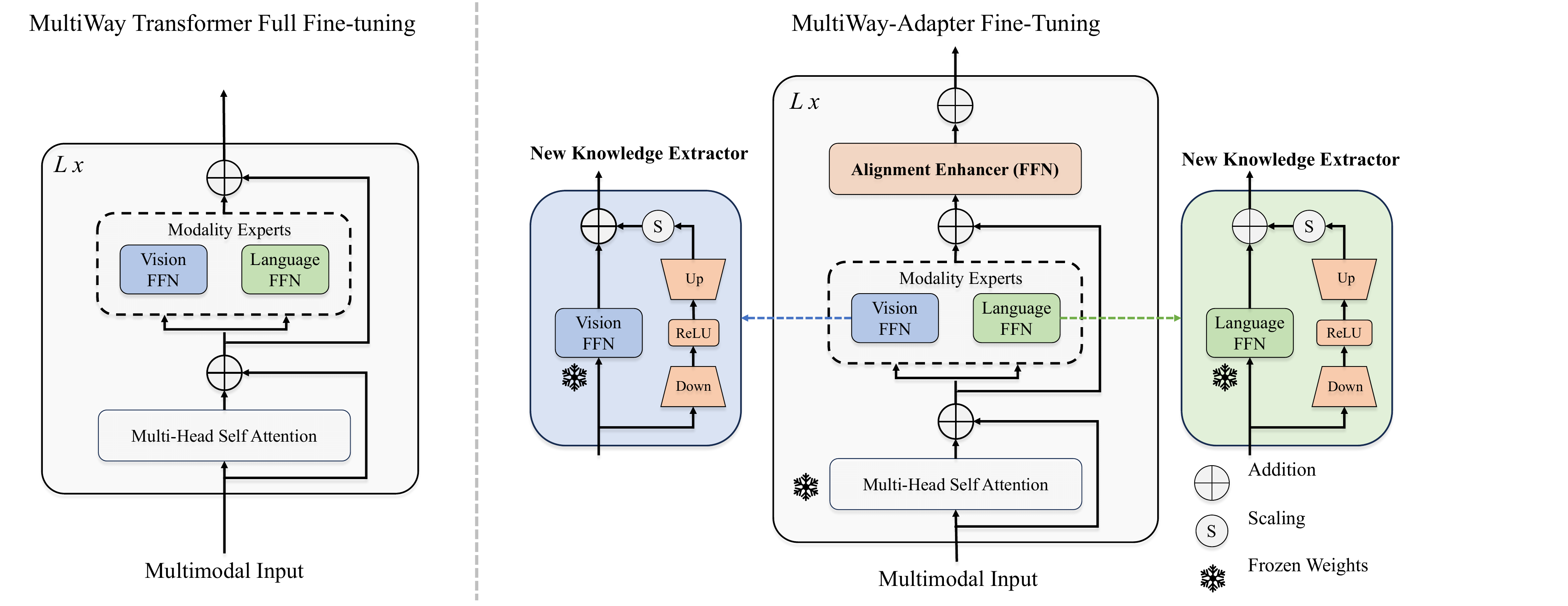}
    \caption{\textbf{Comparison of a MultiWay Transformer and our MultiWay-Adapter fine-tuning.} MultiWay-Adapter uses a dual-component design, including New Knowledge Extractor and Alignment Enhancer. We replace the original FFN with New Knowledge Extractor:  frozen branch (left) and the trainable bottleneck module (right). Moreover, we add a Alignment Enhancer upon the original FFN to enhance the inter-modal alignment.}
    \label{fig:overallarchi}
\end{figure*}

\section{Introduction}
\label{sec:intro}
Recent advancements in Multimodal Large Language Models (MLLMs), such as BLIP2~\cite{DBLP:conf/icml/0008LSH23} and BEiT-3~\cite{beit3}, have demonstrated state-of-the-art performance in multimodal tasks, exemplified by their capabilities in Visual Question Answering. However, the adaptation of these MLLMs to specialized downstream tasks remains a substantial challenge, particularly for image-text retrieval, a common use-case in multimodal learning. Traditional full fine-tuning requires isolated, exhaustive retraining for each new task, demanding intensive computational resources and thus limiting practical applications. For instance, training BLIP2-Giant on an Nvidia A100 GPU takes 144 days~\cite{DBLP:conf/icml/0008LSH23}.

Given the challenge of fine-tuning MLLMs, there is a growing need to develop efficient adaptation methods for MLLMs~\cite{DBLP:conf/iclr/HuSWALWWC22,DBLP:conf/cvpr/Sung0B22}. While progress has been made in unimodal domains using adapter modules, these methods remain largely underexplored in multimodal contexts, particularly for image-text retrieval. Furthermore, existing adaptation methods for MLLMs~\cite{DBLP:conf/cvpr/Sung0B22,zhai2022lit,DBLP:conf/nips/ChenGTWSWL22} focus on information extraction from downstream datasets but neglect the critical need for inter-modal alignment. The goal of inter-modal alignment is to bring different modalities into a common feature space where they can be effectively compared, combined, or related. With shallow alignment, the model would fail to capture the complex inter-relations between different modalities, thereby impacting its effectiveness in multi-modal tasks~\cite{baltruvsaitis2018multimodal,DBLP:conf/iclr/SuZCLLWD20,longmultimodal}.

To address the issue of shallow inter-modal alignment while preserving the efficiency advantages of adapter approaches, we introduce the MultiWay-Adapter (MWA), a lightweight yet effective framework designed explicitly for MLLMs adaptation. Additional components of MWA are small in size but bring a significant performance boost in transfer learning with minimal fine-tuning cost. Our key contributions include:

\begin{itemize}
\vspace{-2mm}
\item We propose MWA that incorporates a dual-component approach, namely the New Knowledge Extractor and the Modality Enhancer. MWA not only extracts new knowledge from downstream datasets but also ensures deep inter-modal alignment, which is crucial for superior performance in vision-language tasks. To the best of our knowledge, this paper is the first work that mitigates the issue of shallow inter-modal alignment in adapter approaches for MLLMs.
\vspace{-2mm}
\item Through comprehensive experiments, we demonstrate that MWA achieves superior zero-shot performance on the Flickr30k dataset by tuning merely an additional $2.58\%$ of parameters to the BEiT-3 Large model, saving up to $57\%$ in fine-tuning time compared to full-model fine-tuning. MWA also demonstrates no statistically significant decreases in performance in other settings, compared to full fine-tuning, requiring significantly fewer resources.
\vspace{-2mm}
\item Experimental results demonstrate the robustness of MWA when parameters scale up, making it ready for MLLMs that are continually increasing in size.
\vspace{-2mm}
\item \looseness -1 Our ablation study confirms the effectiveness of both MWA components, substantiating our design choices.
\end{itemize}

\section{Related Work}
\label{sec:relatedwork}
\vspace{-1mm}
\looseness=-1 \textbf{Challenges in Adapting Large Multimodal Models}.
Recently, increasing model size has been shown to be an effective strategy for improving performance. Models such as BEiT-3~\cite{beit3} and BLIP-2~\cite{DBLP:conf/icml/0008LSH23}, with up to 1.9 billion and 12.1 billion parameters, respectively, have set new state-of-the-art results in multimodal tasks such as Visual Question Answering. However, their application to specialized downstream tasks is often limited by computational constraints~\cite{long2023crisisvit,long2023robollm,yi2023large,long2023elucidating,long2023lacvit,longautocrisis}. For instance, the requirement for large GPU memory in full fine-tuning limits their adaptations for specialized tasks on commodity hardware, e.g., 45GB for full fine-tuning of the BEiT-3 Large model.

\looseness=-1 \vspace{1mm} \noindent \textbf{Efficient Transfer Learning Methods}.
The challenge of computational efficiency in fine-tuning MLLMs has given rise to Parameter-Efficient Transfer Learning (PETL) methods. These are broadly categorized into partial parameter updates~\cite{10095154} and modular additions~\cite{10094795,DBLP:conf/cvpr/Sung0B22}. The former is resource-intensive and model-specific, while the latter adds new modules to architectures, updating only these components. However, most studies only focus on unimodal tasks in domains such as vision~\cite{DBLP:conf/nips/RebuffiBV17}, text~\cite{DBLP:conf/icml/HoulsbyGJMLGAG19} or audio~\cite{DBLP:conf/icassp/ThomasKK22, DBLP:conf/icassp/Eeckth23a, DBLP:conf/icassp/KesslerTK22}, neglecting multimodal tasks. A few works~\cite{DBLP:conf/cvpr/Sung0B22,zhai2022lit,10095154,10094923} target multimodal tasks but suffer from shallow inter-modal alignment. Our work introduces the MultiWay-Adapter, designed for efficient MLLM transfer learning and enhanced inter-modal alignment.

\section{Approach} \label{sec:approach}
\looseness=-1 We introduce MultiWay-Adapter (MWA), designed for the efficient transfer of Multimodal Large Language Models (MLLM) to downstream tasks. Although the primary focus of this paper is on image-text retrieval tasks, the potential applicability of the MWA is broader, such as video text retrieval and image captioning. 

\vspace{1mm} \noindent \textbf{Preliminaries}.
\looseness=-1 The overall framework is constructed on the basis of a popular architecture of MLLM, which utilizes a MultiWay Transformer design~\cite{beit3}. As depicted on the left of Figure~\ref{fig:overallarchi}, each MultiWay Transformer block comprises a shared self-attention module and a pool of feed-forward networks (i.e., modality experts) tailored for different modalities. This design is similar to the dual-backbones architecture of multimodal models, e.g., one encoder for vision input and another encoder for language input, yet differs by sharing the weights within each self-attention module. This design choice reduces the parameter count and enhancing inter-modal alignment—an essential quality for high-performance multimodal tasks~\cite{beit3}.
\vspace{-3mm}
\subsection{MultiWay-Adapter}
\vspace{-1mm}
\textbf{Overall Architecture.} 
Our proposed MWA uses a dual-component approach: the New Knowledge Extractor and the Alignment Enhancer, as illustrated on the right of Figure~\ref{fig:overallarchi}.

\vspace{1mm} \noindent \textbf{New Knowledge Extractor.}
The New Knowledge Extractor is designed for extracting new knowledge from the target downstream tasks. In contrast to the conventional full fine-tuning of MultiWay Transformers, we replace both feed-forward networks (FFNs) in the transformer block with a New Knowledge Extractor. This extractor comprises two branches: the left branch, identical to the original network, and an additional right branch introduced for task-specific fine-tuning. The latter utilizes a bottleneck structure to limit the number of parameters and includes a down-projection layer and an up-projection layer. Formally, for a specific input feature \( {x_{i}}' \), the right branch of the New Knowledge Extractor produces the adapted features, \( \tilde{x_{i}} \), as:
\begin{equation}
\vspace{-1mm}
\tilde{x_{i}} = \text{ReLU}(\text{LN}({x_{i}}') \cdot \textbf{W}_{\text{down}}) \cdot \textbf{W}_{\text{up}}
\vspace{-1mm}
\end{equation}

\begin{table*}[t]
\centering
\resizebox{1\textwidth}{!}{
\begin{tabular}{@{}llccccccc@{}}
\toprule
             &                  &               & \multicolumn{1}{l}{} & \multicolumn{1}{l}{} & \multicolumn{2}{l}{MSCOCO (5k test set)} & \multicolumn{2}{l}{Flickr30k (1k test set)} \\ \midrule
Model        & FT-Way  & Tunable params (M)    & GPU Mem (GB)         & Time (Min)                & IR@1                & TR@1               & IR@1                 & TR@1                 \\ \midrule
ALBEF~\cite{li2021align}        & Full Fine-tune   & 196           & N/A                  & N/A                     & 60.7                & 77.6               & 85.6                 & 95.9                 \\
ALIGN~\cite{jia2021scaling}       & Full Fine-tune   & 825           & N/A                  &  N/A                    & 59.9                & 77.0               & 84.9                 & 95.3                 \\ \hline
BEiT-3-Base  & Full Fine-tune   & 222 (100\%)   & 37GB                 & 225                     & 61.4                & 79.0               & 86.2                 & 96.3                 \\
BEiT-3-Large & Full Fine-tune   & 675 (304\%)   & 45GB                 &   353                   & 63.4 \risedata{(+2.0)}                & 82.1  \risedata{(+3.1)}              & 88.1 \risedata{(+1.9)}                 & 97.2 \risedata{(+0.9)}                 \\
BEiT-3-Base  & MultiWay-Adapter & 7.13 (\textbf{3.21\%}) & \textbf{30GB}                 & \textbf{130}                     & 60.7 \downdata{(-0.7)}                & 78.3   \downdata{(-0.7)}            & 85.4  \downdata{(-0.8)}               & 95.4  \downdata{(-0.9)}               \\
BEiT-3-Large & MultiWay-Adapter & 17.40 (\textbf{2.58\%}) & \textbf{36GB}                 & \textbf{194}                     & 63.3 \risedata{(+1.9)}                    & 82.1  \risedata{(+3.1)}               & 88.0 \risedata{(+1.8)}          & 97.1  \risedata{(+0.8)}                 \\ \bottomrule
\end{tabular}
}
\caption{\textbf{Comparative Analysis of Full Fine-Tuning and the MultiWay-Adapter}: The table shows Top-1 recall metrics on COCO and Flickr30k datasets, presented as both absolute values and relative gaps to the BEiT-3 Base full fine-tuning Model. Metrics for Text-to-Image Retrieval (IR) and Image-to-Text Retrieval (TR) are provided. GPU memory usage and training time are also included. Training time is measured using a single NVIDIA A6000 GPU with 48GB memory for one epoch. }
\label{table:finetune}
\end{table*}

Here, \( \textbf{W}_{\text{down}} \in \mathbb{R}^{d\times \check{d}} \) and \( \textbf{W}_{\text{up}} \in \mathbb{R}^{\check{d}\times d} \) denote the down-projection and up-projection layers, respectively. \( \check{d} \) is the bottleneck middle dimension and satisfies \( \check{d} \ll d \). LN denotes LayerNorm. This bottleneck module is connected to the original FFN (left branch) through a residual connection via a scale factor \( \alpha \). Then, these features, \( {x_{i}}' \) and \( \tilde{x_{i}} \), are fused with the original one, \( x_{i} \), through a residual connection:
\begin{equation}
\vspace{-1mm}
    x_{i} = FFN(LN({x_{i}}')) + \alpha \cdot \tilde{x_{i}} + x_{i}'
\vspace{-1mm}
\end{equation}

\vspace{1mm} \noindent \textbf{Alignment Enhancer.} 
After extracting new knowledge from the target downstream task, to maintain and improve inter-modal alignment, an Alignment Enhancer module is added atop the pool of feed-forward networks. This module mimics the architecture of the New Knowledge Extractor but uses a larger middle dimension to facilitate better feature fusion and alignment.

During the fine-tuning phase, only the parameters of these newly added modules are optimized, while the rest of the model is frozen (as indicated by the frozen sign in Figure~\ref{fig:overallarchi}). This strategy makes MWA a plug-and-play module, applicable to other MLLM, such as CLIP~\cite{radford2021learning}, VLMo~\cite{bao2022vlmo}, and ALIGN~\cite{jia2021scaling}.

\section{Experiments}

\looseness -1 \textbf{Setup}.
We conducted experiments on two state-of-the-art MLLMs, BEiT-3 Base and BEiT-3 Large, across two widely-used image-text retrieval datasets: MSCOCO~\cite{lin2014microsoft} and Flickr30K~\cite{plummer2015flickr30k}. We use the 5k test set of MSCOCO and 1k test set of Flickr30k to report metrics, in accordance with previous studies~\cite{lin2014microsoft,plummer2015flickr30k}. We initialized the backbone, excluding our additional modules, with pre-trained weights, which were frozen during the fine-tuning process when employing MultiWay-Adapter. For fine-tuning, the batch size is 512 for the Large model and 1024 for the Base model, over 20 epochs with an initial learning rate of \(0.001\). Middle dimensions for the New Knowledge Extractor and the Alignment Enhancer were set to 64 and 128, respectively. All the code used in our experiments can be found in \url{https://github.com/longkukuhi/MultiWay-Adapter}.

\vspace{1mm} \noindent \textbf{Experimental Results}.
The objective of this experiment is to assess the efficiency and efficacy of our MWA framework in comparison to traditional full fine-tuning methods. We compared our MWA approach with full fine-tuning in two distinct settings: fine-tuning performance and zero-shot performance. \looseness=-1

\vspace{1mm} \noindent \textbf{Fine-Tuning Performance}: As shown in Table \ref{table:finetune}, our MWA method demonstrates superior computational efficiency. Specifically, it utilizes a mere \(3.21\%\) and \(2.58\%\) of the trainable parameters for the Base and Large variants of BEiT-3, respectively, in contrast to conventional full fine-tuning. This leads to a substantial reduction in GPU memory consumption—by 7GB and 9GB for the Base and Large variants, respectively. Furthermore, MWA significantly reduces the time required for fine-tuning. For instance, fine-tuning MWA with the BEiT-3 Base model is reduced by \(57\%\) compared to full fine-tuning. \looseness=-1

Regarding effectiveness, the performance decrement when utilizing MWA is statistically insignificant for both the Base and Large BEiT-3 variants, with deviations falling within a margin of less than $1\%$. Synthesizing these efficiency and effectiveness attributes demonstrates that MWA, when applied to the BEiT-3 Large model, consumes merely \(86\%\) of the time required for full fine-tuning of the BEiT-3 Base model, yet surpasses its performance. This suggests that MWA enables enhanced performance with reduced computational time, particularly for larger models. Additionally, as the model size increases, the performance disparity between MWA and full fine-tuning diminishes, indicating a positive correlation between MWA's effectiveness and model size. \looseness=-1

\vspace{1mm} \noindent \textbf{Zero-Shot Performance}: To evaluate the transfer capabilities of MWA and full fine-tuned methods, we conducted experiments in a zero-shot setting. In this setting, the model is evaluated on Flickr30k (1k test set), with which it has no prior knowledge of, thereby necessitating reliance on intrinsically learned knowledge to simulate the handling of previously unseen samples. These models were initially fine-tuned on the MSCOCO dataset. As shown in Table \ref{table:zeroshot}, MWA surpasses the performance of full fine-tuning when employed with the BEiT-3 Large model. We hypothesize that this enhancement is attributable to the preservation of generalizable knowledge in the frozen weights, knowledge potentially lost during the full fine-tuning process. This retained knowledge augments the model's ability to adeptly manage unseen instances. Thus, MWA not only match the performance of full fine-tuning method but also distinguishes itself in terms of resource efficiency and transferability. \looseness=-1

In summary, the experimental results demonstrate that MWA serves as an effective and resource-efficient fine-tuning method for MLLMs, especially when computational resources are constrained. \looseness=-1

\begin{table}[]
\centering
\resizebox{0.42\textwidth}{!} {
\begin{tabular}{@{}llcc@{}}
\toprule
        &            & \multicolumn{2}{c}{Flickr30k} \\ \midrule
Model        & FT-Way           & IR@1  & TR@1  \\ \midrule
BEiT-3-Large & Full fine-tune   & 85.99 & 95.48 \\
BEiT-3-Large & MultiWay-Adapter & 86.26 & 95.51 \\ \bottomrule
\end{tabular}
}
\caption{\textbf{Zero-shot performance on Flickr30k}. }
\label{table:zeroshot}
\vspace{-1mm}
\end{table}

\section{Analysis}

\begin{figure}
\centering

\includegraphics[height=0.6\linewidth]{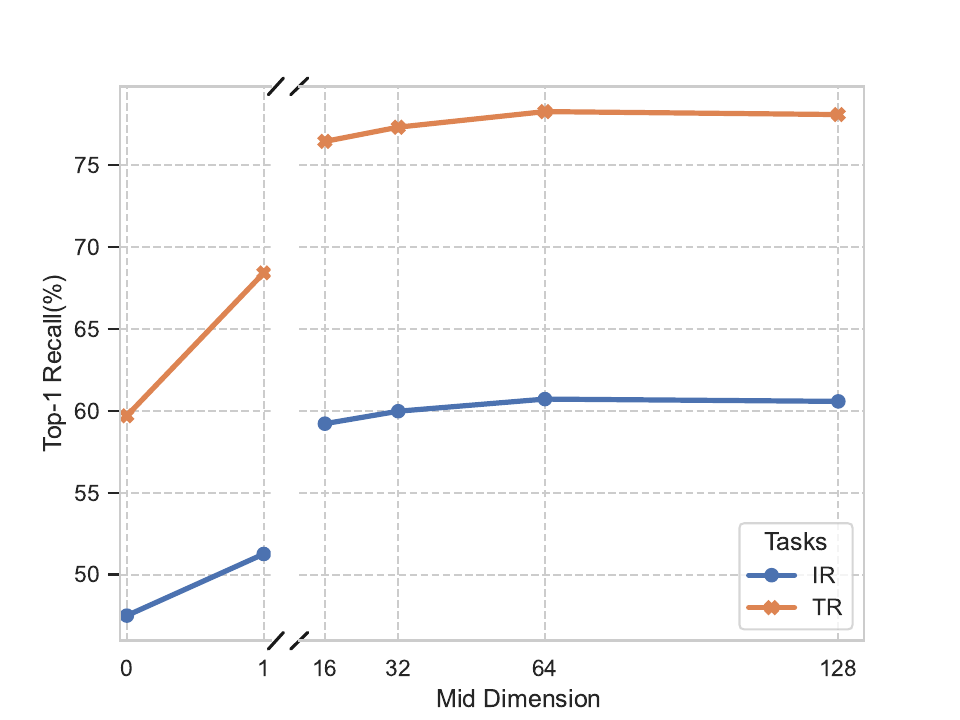}
\caption{\textbf{Evaluation of different sizes of mid-dimension New Knowledge Extractor on MSCOCO}.}
\label{fig:dim}
\vspace{-2mm}
\end{figure}

\noindent \textbf{Scaling Tunable Parameters Up}:  
\looseness=-1 The primary aim of this section is to investigate the impact of varying the number of tunable parameters on performance and to identify the optimal value for additional parameters. The ``mid-dimension" of the New Knowledge Extractor largely controls the number of tunable parameters. We conducted an empirical evaluation across a range of mid dimensions \{0, 1, 16, 32, 64, 128\}  on the MSCOCO dataset using the BEiT-3 Base model. The results are summarized in Figure~\ref{fig:dim}. The data reveals a noticeable increase in performance as the dimension grows, plateauing at 64. Specifically, we observed a peak performance gain of \(9.45\%\), in text to image retrieval when increasing the dimension from 1 to 64. This indicates that increasing the number of parameters in the adapter does not guarantee performance improvement. When the dimension is set to zero, it represents the zero-shot performance of the BEiT-3 Base model without MWA. Notably, MWA delivers superior performance compared to the zero-shot performance of the BEiT-3 Base model, even when the mid-dimension is as low as one. Furthermore, performance variability is relatively small when increasing the dimension from 16 to 64, indicating that MWA is stable in tuning and not sensitive to changes in size.

\noindent \textbf{Ablation on MultiWay Adapter's Components}:  
\looseness=-1 In this section, our focus is to quantify the individual contributions of our two newly introduced components: the New Knowledge Extractor and the Alignment Enhancer. An ablation study was performed on the MSCOCO dataset using the BEiT-3 Base model. The performance metrics for each component, both in isolation or in combination, are detailed in Table~\ref{table:ablation}. Our findings demonstrate that omitting either component leads to a significant decline in performance, approximately \(3\%\), for image to text retrieval and around \(4\%\), for text to image retrieval. Importantly, the Alignment Enhancer, a novel element distinct from previous Adapter methods, validates its critical role in maintaining deep alignment between modalities through observed performance gains. In summary, both components not only significantly contribute to the overall performance but also complement each other effectively.

\begin{table}[]
\centering
\resizebox{0.28\textwidth}{!} {
\begin{tabular}{@{}lcccc@{}}
\toprule
     &  &    & \multicolumn{2}{c}{MSCOCO} \\ \midrule
Model & KE   & AE   & IR@1                & TR@1               \\ \midrule
BEiT-Base &  &      & 61.40              & 79.00              \\
BEiT-Base & \checkmark &      & 57.32               & 73.92              \\
BEiT-Base &     & \checkmark & 57.88               & 74.61              \\
BEiT-Base  & \checkmark & \checkmark & 60.72               & 78.26              \\ \bottomrule
\end{tabular}
}
\caption{\textbf{Ablation study of two modules of MultiWay-Adapter}. KE refers to the New Knowledge Extractor and AE refers to the Alignment Enhancer. }
\vspace{-4mm}
\label{table:ablation}
\end{table}

\vspace{-1mm}
\section{Conclusion}
\vspace{-1mm}
\looseness=-1 We introduce the MultiWay-Adapter (MWA), an effective framework designed for the efficient adaptation of Multimodal Large Language Models (MLLM) to downstream tasks. Addressing the issue of shallow inter-modal alignment in existing methods, MWA employs a dual-component approach, utilizing both the New Knowledge Extractor and the Alignment Enhancer. This strategy enables MWA to not only extract novel information from downstream datasets but also to secure deep inter-modal alignment. Our empirical findings reveal that with the addition of a mere \(2.58\%\) in extra parameters, there is no statistically significant decline in performance across all tested settings while reducing the fine-tuning time by up to \(57\%\). Our research paves the way for future studies on efficient multimodal fine-tuning methods and holds potential for extension into other vision-language tasks.

\clearpage
\bibliographystyle{IEEEbib}
\bibliography{strings,refs}

\end{document}